\documentclass[final]{cvpr}

\usepackage{wrapfig}
\usepackage{times}
\usepackage{epsfig}
\usepackage{graphicx}
\usepackage{amsmath}
\usepackage{amssymb}
\usepackage{multirow}
\usepackage[numbers,sort]{natbib}
\usepackage{algorithmic}
\usepackage{algorithm}
\usepackage{cancel}
\usepackage{booktabs}

\usepackage{qiangstyle}
\usepackage{setspace}

\usepackage{graphicx}
\usepackage{tabu}
\usepackage{multirow}
\usepackage[pagebackref=true,breaklinks=true,colorlinks,bookmarks=false]{hyperref}

\renewcommand{\L}{{\mathcal L}}

\def\cvprPaperID{7379} 
\def\confYear{CVPR 2021}

\begin{document}
\title{AlphaMatch: 
Improving Consistency for Semi-supervised Learning with Alpha-divergence }

\author{
Chengyue Gong, Dilin Wang, Qiang Liu \\
\\
University of Texas, Austin 
}

\maketitle


\begin{abstract}
Semi-supervised learning (SSL) is a key approach toward more data-efficient machine learning by jointly leverage both labeled and unlabeled data. 
We propose \emph{AlphaMatch}, an efficient SSL method 
that leverages data augmentations, by efficiently enforcing the label consistency between the data points and the augmented data derived from them.  
Our key technical contribution lies on: 1) using alpha-divergence to prioritize the regularization on data with high confidence, 
achieving similar effect as FixMatch \citep{sohn2020fixmatch} but in a more flexible fashion, and 2) proposing an optimization-based, EM-like algorithm to enforce the consistency, 
which enjoys better convergence than iterative regularization procedures used in recent SSL methods such as FixMatch, UDA, and MixMatch. 
AlphaMatch is simple and easy to implement, 
and consistently outperforms prior arts on standard benchmarks, e.g. CIFAR-10, SVHN, CIFAR-100, STL-10. Specifically, we achieve 91.3\% test accuracy on CIFAR-10 with just 4 labelled data per class, 
substantially improving over the previously best 88.7\% accuracy achieved by FixMatch.


\end{abstract}

\section{Introduction}
Semi-supervised learning (SSL) \citep{chapelle2009semi}
is a powerful paradigm for leveraging both labeled and unlabeled data jointly in machine learning (ML). 
%
Effective SSL methods can help build accurate prediction out of very limited labeled data,
and can also boost state-of-the-art performance and robustness in typical supervised learning by leveraging extra unlabeled data  \citep[see e.g.,][]{xie2019self, kahn2020self, zhai2019adversarially}. 
Due to the high cost of collecting labels and 
the availability of a vast amount of unlabeled data, 
breakthroughs in SSL can dramatically 
advance the application of ML in countless fields. 

Recently, data augmentation has been shown a powerful tool for developing  state-of-the-art SSL methods,  
including unsupervised data augmentation (UDA) \citep{xie2019uda},  FixMatch \citep{sohn2020fixmatch}, MixMatch \citep{berthelot2019mixmatch}, ReMixMatch \citep{berthelot2019remixmatch} and $\Pi$-model \citep{rasmus2015pi}. 
All these methods are  based on 
the similar idea of enforcing the consistency between 
the label of a data point and that of its perturbed version generated by data augmentation. 
This encourages the learned models to be invariant under given data augmentation transforms, 
hence incorporating 
inherent
structures of the data into semi-supervised learning. 

The performance of these algorithms can be critically influenced by 
what matching objective and matching algorithm are used to enforce the consistency. 
For the objective, 
UDA applies a KL divergence penalty to enforce the consistency uniformly on all the data points.
More recently, FixMatch shows that it is useful to 
focus on matching the consistency on the high confidence data points with a \emph{hard thresholding}  approach, 
by applying regularization only on data with  confidence higher than a threshold 
and use the hard label as the target.  
%
In terms of the matching algorithm, 
most of the existing methods, including UDA and FixMatch, 
are based on a similar iterative regularization procedure that uses the label distribution predicted from the previous iteration as the target for the next step.
Although being simple and intuitive, 
a key problem is that this iterative procedure does not correspond to optimizing a fixed objective function, and hence may suffer from instability and non-convergence issues.  

This work proposes two key algorithmic advances to improve the objective and algorithm for consistency matching in SSL:  
 1) we propose to use alpha-divergence to measure the label consistency. 
We show that, by using a large value of $\alpha$ in alpha-divergence, we can focus more on high confidence 
instances in a way 
similar to the hard-thresholded regularization of FixMatch, but in a more ``soft'' and flexible fashion.
2) We propose an optimization-based framework for consistency matching, which yields an  
EM-like algorithm 
with better convergence property than the commonly used iterative regularization procedures. 
By combining these two key techniques, 
our main algorithm \emph{AlphaMatch} 
yields better SSL with more effective and stable consistency regularization. 

Empirically, 
we find that AlphaMatch consistently outperforms 
recently-proposed SSL methods such as FixMatch, ReMixMatch, MixMatch, and UDA  both in terms of  accuracy and data efficiency, 
on various benchmarks including  CIFAR-10, SVHN CIAFR-100, and STL-10.  
In particular, our method improves over the state-of-the-art method, FixMatch, across all the settings we tested. 
Our improvement is particularly significant  
when the labels are highly limited. 
For example, on CIFAR-10, 
we improve the 88.71\%$\pm$3.35\% accuracy of FixMatch to 91.35\%$\pm$3.38\% 
when only 4 labelled images per class are given.  

\section{Background: Semi-Supervision with Data Augmentation
}
\label{sec:bg}
%
We give a brief introduction to 
unsupervised data augmentation (UDA) \citep{xie2019uda} and FixMatch \citep{sohn2020fixmatch}, which are mostly related to our work. 
Denote by $\D_s$ and $\D_u$ the labeled and unlabeled datasets, respectively.
For a data point $x$ in $\D_u$, let $\P_x$ be 
a distribution that prescribes  
a random perturbation or augmentation transformation on $x$ that (with high probability) keeps the label of $x$ invariant,
such as rotation, shift, and cutout \citep{devries2017cutout}.
For learning a prediction model $p_\theta(y|x)$, 
UDA works by iteratively updating  the parameter $\theta$ via  
\begin{align} 
\begin{split} 
& \theta_{t+1}  \gets \argmin_{\theta} 
\bigg\{ ~\L(\theta; ~\data_{s})  + 
\lambda  \Phi(\theta;\theta_{t},\D_u)~ \bigg\} ~~\text{with~~~~} \\
& \Phi(\theta;~\theta_{t}, \D_u) = \E_{x\sim \data_u, ~ x' \sim \P_x} \bigg [\KL(p_{\theta_t}(\cdot~|~x) ~\parallel~ p_\theta(\cdot ~|~ x')) \bigg] , 
\end{split}
\label{eq:ssl-loss}
\end{align}
where $\theta_t$ denotes the value at the $t$-th iteration,  
$\L(\data_{s};~ \theta) $ is the typical supervised loss, e.g. the cross entropy loss, 
and  
$\KL(\cdot ~\pp~\cdot)$ denotes the Kullback–Leibler (KL) divergence. 
Here  $\Phi(\theta;\theta_{t},\D_u)$  
can be viewed as a consistency regularization 
that enforces the label distribution  of the augmented data $x'\sim \P_x$ to be similar to that of the original data $x$ (based on the parameter $\theta_t$ at the previous iteration); $\lambda$ is a regularization coefficient. 
%

In practice, the optimization in \eqref{eq:ssl-loss}
can be approximated by applying one step of gradient descent initialized from $\theta_t$, yielding 
\begin{align} 
\theta_{t+1} \gets 
\theta_{t} -  
\epsilon \nabla_{{\theta}}\bigg (\L(\theta; ~\data_{s})  + \lambda \Phi( \theta; ~\theta_{t}, ~ \D_u)
\bigg) \bigg|_{\theta = \theta_{t}},  \label{equ:udagrad}
\end{align}
where $\epsilon$ is the step size. 
This procedure is closely related to  VAT \citep{miyato2015vat}, in which augmented data $x'$ is replaced by an adversarial example in a neighboring ball of $x$.
%

FixMatch improves UDA with ideas similar to the classical \emph{pseudo-labeling} method \citep{grandvalet2005semientropy}.
FixMatch replaces the ``soft label'' $p_{\theta_t}(\cdot|x)$ 
with the corresponding ``hard label''
$\hat y_{\theta_t}(x) = \argmax_y p_{\theta_t}(y~|~x)$ (a.k.a. pseudo-label), and  turns on the regularization only when   
the confidence of the pseudo-label, estimated by  $p_{\theta_t}(\hat y_{\theta_t}(x)~|~x)$, is sufficiently large: 

\begin{align} 
\begin{split}
&\Phi(\theta; ~\theta_t, \D_u) := \E_{x\sim \D_u, x'\sim \P_x} \bigg [\ind\bigg(p_{\theta_t}(\hat y_{\theta_t}(x)~|~x) \geq \tau\bigg) \\ & \times \KL \bigg(\hat p_{\theta_{t}}(\cdot~|~x) ~\pp~ p_\theta(\cdot ~|~ x') \bigg) \bigg], \label{equ:fixmatch}
\end{split}
\end{align}

where $\hat p_{\theta_{t}}(y~|~x) := \delta(y = \hat y_{\theta_t}(x))$ and $\ind(\cdot)$ is the indicator function and $\tau$ a threshold parameter (e.g., $\tau = 0.95$). 
This regularization has  two important effects:
1) it upweights the hard label $\hat y_{\theta_t}(x)$ while discarding all the other labels from the regularization, 
and 2) it skips the data points 
with low confidence (i.e., $p_{\theta_t}(\hat y_{\theta_t}(x)~|~x) < \tau$).

\paragraph{Remark} 
Note that
the iterative procedure in \eqref{eq:ssl-loss}-\eqref{equ:udagrad}  
does not in general correspond  to optimizing a fixed objective function, 
and hence does not guarantee to converge theoretically and may suffer from non-convergence practically. 
For example, we empirically observe that the performance of UDA tends to degenerate significantly when trained for many iterations when few labelled data is given, and FixMatch is sensitive to the choice of threshold $\tau$. 
An alternative is to directly optimize the following objective function: 
\begin{align} \label{equ:phithth}
\min_{\theta} \bigg\{ \L(\theta; ~ \data_s)  + \lambda \Phi(\theta; ~ \theta, ~ \data_u) \bigg\},
\end{align}
whose gradient descent yields 
\begin{align} 
\theta_{t+1} \gets 
\theta_{t} - 
\epsilon \nabla_{{\theta}}\bigg (\L(\theta; ~\data_{s})  + \lambda \Phi( \theta; ~\theta, ~ \D_u) 
\bigg) \bigg|_{\theta = \theta_{t}}.  \label{equ:ggrad}
\end{align}
The difference with \eqref{equ:udagrad} is that the gradient of $\Phi(\theta; ~ \theta_t, \data_u)$ through $\theta_t$ 
 is detached and dropped in  \eqref{equ:udagrad}, 
while the gradient of $\Phi( \theta; ~\theta, ~ \D_u) $ in \eqref{equ:ggrad} needs to be taken for both $\theta$. In \citet{miyato2015vat, xie2019uda, sohn2020fixmatch} and all the related works, \eqref{equ:udagrad} is chosen over \eqref{equ:ggrad} for better empirical performance, likely because stopping the gradient encourages the information to pass from the clean data $x$ to the augmented data $x'$, but not the other way,  so that the supervised objective is less interfered by the consistency regularization than in the direction optimization approach \eqref{equ:ggrad}. 
A complete theoretical understanding of the benefit of stopping gradient is still an open question. 

\begin{figure*}[h]
\centering
\setlength{\tabcolsep}{1pt}
\includegraphics[height=0.32\textwidth]{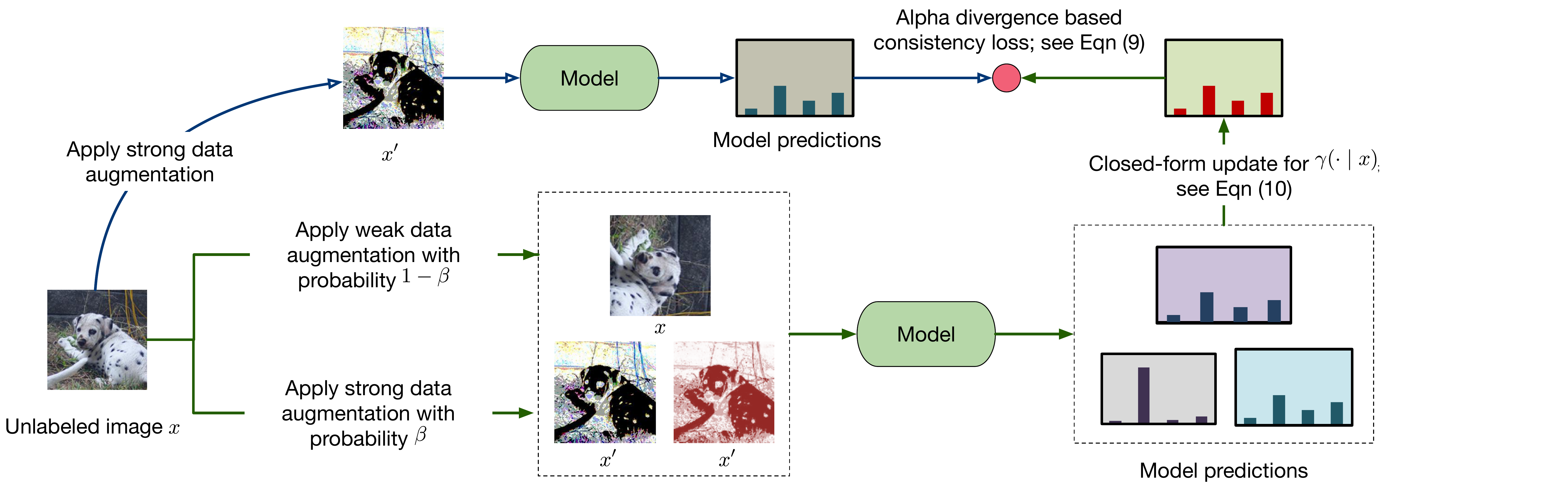} 
\caption{
Diagram of our proposed semi-supervised learning algorithm. When augmentation, we use a hyper-parameter $\beta$ to control a mixture of strong and weak augmentation. When calculating the loss, we use a EM-style update. 1) Solving a closed-from solution of averaging logits with alpha divergence. 2) Using alpha divergence to construct a label-consistency loss. Images are from ImageNet \cite{deng2009imagenet}. 
}
\label{fig:diagram}
\end{figure*}

\section{Our Method}

We introduce our main method \emph{AlphaMatch} (see Algorithm~\ref{alg:main}),  
which consists of two key ideas: 
i) 
we leverage alpha-divergence to  enforce the label consistency  
between augmented and original data in SSL, which can benefit from focusing more on high confidence data  like FixMatch, 
but in a more smooth and flexible fashion (Section~\ref{sec:alpha});  
ii) we introduce 
a convergent 
EM-like algorithm based on an optimization-based framework to replace the iterative procedure in \eqref{eq:ssl-loss}-\eqref{equ:udagrad}, which allows us to enforce the label consistency much efficiently.

\subsection{Matching with Alpha Divergence}
\label{sec:alpha}
%
We propose to use alpha-divergence in 
SSL consistency matching. 
Under the general framework of \eqref{eq:ssl-loss}, this amounts to replace the consistency regularization with 
\begin{align*} 
\Phi(\theta; ~\theta_t, \mathcal D_u) 
= \E_{x\sim \D_u, x'\sim \P_x} \bigg[D_\alpha(p_{\theta_t}(\cdot ~|~ x) \pp~ p_\theta(\cdot ~|~ x'))\bigg], 
\end{align*} 
where $D_\alpha(\cdot ~\pp~ \cdot)$ is the alpha divergence with  $\alpha\in (0,1) \cup (1,\infty)$, defined as: 
\begin{align*} 
\begin{split} 
& D_\alpha(p_{\theta_t}(\cdot|x)~||~p_\theta(\cdot|x'))
\\ & = \frac{1}{\alpha(\alpha-1)}\left (\E_{y\sim p_{\theta_t}(\cdot|x)} \left[ \rho_{D_\alpha}(y|x)\right]
-1 \right), \\ &
~~\text{with}~~\rho_{D_\alpha}(y|x) := \bigg({\frac{p_{\theta_t}(y|x)}{p_{\theta}(y|x')} \bigg)^{{\alpha-1}}}.
\end{split}
\end{align*}
It is well-known that $D_\alpha(\cdot||\cdot)$ reduces to KL divergence when $\alpha \to 0$ or $1$, as follows,
\begin{align*} 
& \lim_{\alpha \to 1} D_\alpha(p_{\theta_t}(\cdot|x)~\pp~p_\theta(\cdot|x'))  = \KL(p_{\theta_t}(\cdot|x)~\pp~p_\theta(\cdot|x')), \\
& \lim_{\alpha \to 0} D_\alpha(p_{\theta_t}(\cdot|x)~\pp~p_\theta(\cdot|x'))  = \KL(p_\theta(\cdot|x')~\pp~p_{\theta_t}(\cdot|x)).  
\end{align*}
In general, the value of $\alpha$ critically influences the result of the algorithm. 
The regime of $\alpha > 1$ is  of particular interest for our purpose, 
because it allows us to achieve a FixMatch-like effect but in a ``soft way''. 
This is because when $\alpha$ is large, 
the power term 
$\rho_{D_\alpha}(y|x)$ in $D_\alpha(\cdot||\cdot)$  tends to put a higher weight on the $(x,y)$ pairs with large values  of $p_{\theta_t}(y|x)$, and hence upweighting the importance 
the instances $x$ with high confidence as well as their dominating labels $y$. 
This is similar to what FixMatch attempts to achieve in \eqref{equ:fixmatch}, except that the regularization is enforced in a different and more ``soft'' 
fashion, so that the instances with lower confidence and the less dominant labels still contribute to the loss, except with a lower degree.  

It is useful to get further insights by examining the gradient of $D_\alpha(\cdot||\cdot)$, which equals

\begin{align}
\begin{split}
\!\!\!\!
& \nabla_\theta 
 D_\alpha(p_{\theta_t}(\cdot|x)~\pp~p_\theta (\cdot|x'))  
\\ & =-\frac{1}{\alpha}\E_{y\sim p_{\theta_t}(\cdot | x)}  \bigg[\blue{\rho_{D_\alpha}(y|x)} \nabla_{\theta} \log p_\theta (y|x') \bigg]. 
\label{equ:alpahgrad}
\end{split}
\end{align}
When $\alpha=1$ (corresponding to UDA),  we have $\rho_{D_\alpha}(y|x) = 1$.  
The gradient of FixMatch is also similar but with $\rho_{D_\alpha}(y|x)$ replaced by 
\begin{align}
\begin{split}
& \rho_{\mathrm{FixMatch}} (y|x) = \\ 
& ~~~~\ind\Big(\max_{y'} p_{\theta_t}(y'|x) \geq \tau  \&y ~= \arg\max_{y'}(p_{\theta_t} (y'|x))  \Big).
\end{split}
\end{align}
We can again see that both $\rho_{D_\alpha}(y|x)$ and $\rho_{\mathrm{FixMatch}}(y|x)$ favor the  data and label $(x,y)$ with high  confidence   $p_{\theta_t}(y|x)$, but 
$\rho_{\mathrm{FixMatch}}(y|x)$ does it in a more aggressive fashion. 
In addition, note that $\rho_{D_\alpha}(y|x)$ depends on both $p_{\theta_t}(y|x)$ and $p_{\theta}(y|x)$, while $\rho_{\mathrm{FixMatch}}(y|x)$ only depends on $p_{\theta_t}(y|x)$. 

\paragraph{Remark}
Alpha divergence provides a general framework for distribution matching. 
It's more flexible compared to other divergences, e.g., Jensen–Shannon divergence. By choosing different values of $\alpha$, our alpha divergence generalizes a number of well-known SSL approaches, including UDA ($\alpha=1$) and FixMatch ($\alpha\rightarrow \infty$). 
Furthermore, note that alpha divergence is proportional to the $\alpha$-th moment of density 
ratio $p(\cdot|x)/p(\cdot|x')$ ($x$ clean image and $x'$ augmented image). 
Therefore, when $\alpha$ is large and positive, large $p(\cdot|x)/p(\cdot|x')$ ratio is strongly penalized, preventing the case of $p(\cdot|x') \ll p(\cdot|x)$. 
This offers a natural and flexible way to propagate high confidence label on $x$ to low-confidence examples $x'$. 

\subsection{An EM-like Optimization Framework for Label Matching}

\begin{figure} 
\centering
\begin{tabular}{c}
\raisebox{3.0em}{\rotatebox{90}{\large Test Accuracy}}
\includegraphics[height=0.26\textwidth]{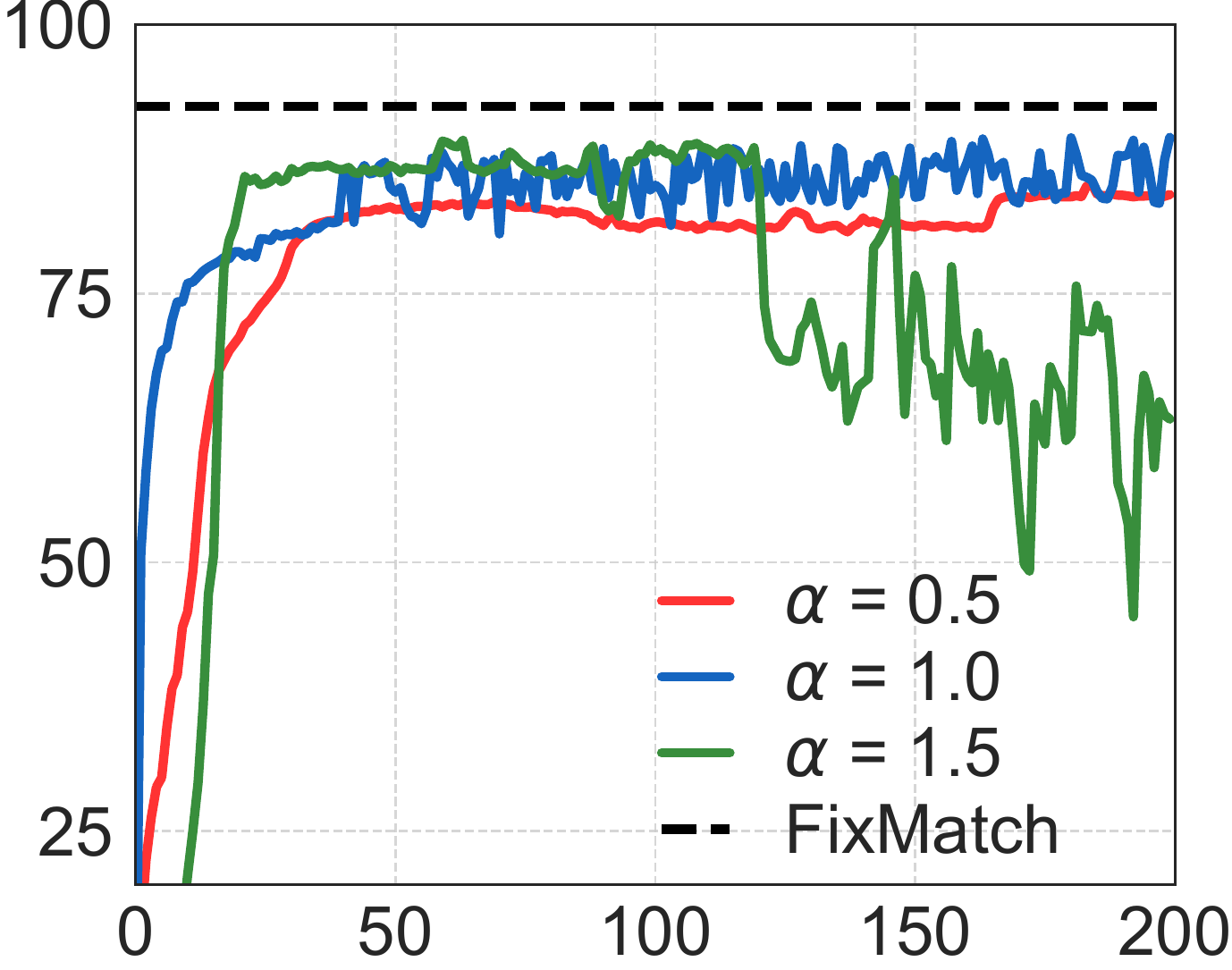} \\
\large{~~~~Training epoch} \\
\end{tabular}
\caption{The learning curve of update Eqn  \eqref{equ:udagrad} with alpha-divergence  with different $\alpha$ 
  on SVHN. 
  The x-axis denotes the training epoch y-axis is the test accuracy. 
  The labelled data includes 4 randomly picked mages per class. 
  The dashed line shows the result of FixMatch. 
  We follow all training settings in FixMatch.
}
\label{fig:large_alpha}
\end{figure}

Despite the attractive property of alpha-divergence, 
we observe that directly incorporating alpha-divergence 
into the iterative update in \eqref{equ:udagrad} like UDA and FixMatch 
tends to cause instability in convergence when $\alpha$ is large, 
which is the regime of main interest. 
We illustrate this with an experiment on SVHN shown in Figure~\ref{fig:large_alpha}, 
which shows that using large $\alpha$ (e.g., $\alpha = 1.5$)
can potentially obtain better results than smaller $\alpha$ 
but is much more unstable during training and may eventually diverge to worse results (note that the case of $\alpha = 1$ (blue curve) is UDA). 
This is because the iterative update in \eqref{eq:ssl-loss} does not  correspond to optimizing a well-defined objective and does not guarantee to converge theoretically.



To address this problem, 
we provide an optimization-based  framework for consistency regularization, 
which yields an EM-like algorithm when solved with a coordinate descent procedure.  
Our method enjoys better convergence and allows us to achieve better performance than directly combining alpha-divergence with \eqref{equ:udagrad}.  
%
Our loss function is 
\begin{align} 
\min_{\theta} \bigg\{ \L(\data_s;~ \theta) + \lambda  
\E_{x\sim \D_u}\bigg[
\min_{\gamma(\cdot~|~x)} 
\Psi_{\alpha,\beta}(\theta, \gamma, x)\bigg] \bigg\}, 
\label{equ:beta_obj}
\end{align}
where $\Psi_{\alpha,\beta}(\theta, x)$ is a consistency regularization on $x$ that we define as  
\begin{align} 
\begin{split} 
& \Psi_{\alpha,\beta}(\theta, \gamma, x) 
= 
(1-\beta ) D_\alpha (\gamma(\cdot~|~x) ~||~ p_\theta(\cdot ~|~x)) \\ & + \beta \E_{x'\sim \P_x}\left [ D_\alpha (\gamma(\cdot~|~x) ~||~ p_\theta(\cdot ~|~x')) \right]   \\
& =  
\E_{x'\sim \P_x^\beta}\left [D_\alpha (\gamma(\cdot~|~x) ~||~ p_\theta(\cdot ~|~x')) \right], 
\end{split}
\label{equ:beta_consistency}
\end{align}

where $\P_x^\beta(x') \overset{\mathrm{def}}{=} (1-\beta) \delta(x'= x) + \beta \P_x(x')$ is a mixture of the random perturbation $\P_x$ and the original data $x$  and $\beta \in [0,1]$ 
is  the ratio between the augmented and original data in $\P_x^\beta$. 
Here $\gamma(\cdot~|~x)$ is an auxiliary variable 
optimized in the space of all distributions. It is 
introduced to serves as a bridge for comparing $p_\theta(\cdot | x)$ and $p_\theta(\cdot | x')$, without having $\theta$ appearing on sides of the divergence like $\Phi(\theta; \theta, \mathcal D)$ in \eqref{equ:phithth}. 
When $\beta = 0.5$, the regularization in \eqref{equ:beta_obj} 
  is a symmetrized version of alpha-divergence that generalizes Jensen-Shannon divergence. 

We optimize our objective function in \eqref{equ:beta_obj} by alterenaively optimizing $\theta$ and $\gamma$: 

\paragraph{Updating $\gamma$}
With  $\theta = \theta_{t}$ fixed,  we update $\gamma(\cdot|x)$ for each $x$: 
\begin{align}
{\gamma_{t}}(\cdot ~|~ x) \gets \argmin_{\gamma(\cdot|x)} 
 \E_{x' \sim \P_x^\beta} \Big [D_\alpha (\gamma(\cdot~|~x) ~\pp~ p_{\theta_{t}}(\cdot ~|~ x'))  \Big]. 
 \label{equ:update_gamma}
 \end{align}

\paragraph{Updating $\theta$} 
With $\gamma = \gamma_{t}$ fixed,  
we update $\theta$ 
by performing gradient descent on 
\begin{align}
 {\theta_{t+1} }
 \gets  \argmin_{\theta} 
 L(\data_{s};~ \theta)  + 
 \lambda \E_{x\sim \D_u} \big [
\Psi_{\alpha,\beta}(\theta, \gamma, x)\big]. 
\label{equ:update_theta}
\end{align}
 Critically, for alpha-divergence, 
 the optimal $\gamma_{t}$ in Eqn~\eqref{equ:update_gamma} 
 equals a simple powered expectation of $p_{\theta_{t}}(\cdot | x')$ as $x'\sim \P_x^\beta$, 
\begin{align} 
\begin{split} 
\gamma_{t}&(\cdot ~|~x)   
\propto \left  (\E_{x'\sim  \P_x^\beta}\left [ p_{\theta_{t}}(\cdot~|~x')^{1 - \alpha}\right]\right )^{\frac{1}{1-\alpha}} =\\
&\bigg(  (1-\beta) p_{\theta_{t}} (\cdot~|~x)^{1-\alpha} +  \beta\E_{x'\sim \P_x}\left [ p_{\theta_{t}}(\cdot~|~x')^{1-\alpha}\right]\bigg )^{\frac{1}{1-\alpha}}.  \\
\end{split}
\label{equ:gammat}
\end{align}

\begin{proof}
Eqn~\eqref{equ:update_gamma} could be viewed as a general optimization of form:
$$
\min_{\gamma } \sum_{i} w_i [D_\alpha(\gamma ~||~p_i) ],
$$
where $w_i$ denotes the importance score for each distribution $p_i$.
With simple calculations, we have, 
\begin{align*}
&\sum_{i} w_i [D_\alpha(\gamma ~||~p_i) ]  = \sum_i w_i \sum_x \frac{\gamma^\alpha(x)}{p_i^{\alpha-1}(x)} \\
&  = \sum_x \sum_i w_i p_i^{1-\alpha} (x) \gamma^\alpha(x),  \\
&= \sum_x \bar{p}^{1-\alpha}(x) \gamma^\alpha(x), \text{with}~ \bar{p}(x) = \bigg(\sum_i w_i p_i^{1-\alpha}(x)\bigg)^{\frac{1}{1-\alpha}} \\ 
&= D_\alpha(\gamma || z \bar{p}), 
\end{align*}
where $z$ is the normalization constant. Therefore, we have
$\gamma^* \propto \bar{p}$.
\end{proof}

This formulation reduces to the typical averaging when $\alpha \to 0$, and 
the reduces to geometric mean when $\alpha \to 1$. 
On the other hand, 
when $\alpha \geq 1$, 
the regime of our interest, 
 $\gamma_{t}(\cdot | x)$ becomes a (powered) harmonic mean of $p_{\theta_{t}}(\cdot | x')$ when $x'\sim \P_x^\beta$. 
Consider the limit $\alpha \to+\infty$, in which case we have $\gamma_{t}(y | x) \approx \mathrm{ess}\inf_{x'\sim \P_x^\beta} p_{\theta_{t}}(y ~|~ x')$, 
which suggests that $\gamma_{t}(y | x)$ is large only when $p_{\theta_{t}}(y ~|~ x')$ are large for both the original data $x'=x$ and  all the augmented data $x'\sim \P_x$.  
This is again consistent with the idea of emphasizing 
high confidence instances. 



\paragraph{Our method}
Combining these two updates yields a simple and practical algorithm shown in Algorithm~\ref{alg:main},
in which we apply one-step of mini-batch gradient descent to update $\theta$ at each iteration and  approximate $\E_{x'\sim \P_x}[\cdot]$ in \eqref{equ:beta_consistency} and \eqref{equ:gammat} 
by drawing a number of $n$ random samples $\{x_i'\}_{i=1}^n$ from $\P_x$ for practical efficiency.  

Our method converges theoretically because it is a coordinate descent by design. 
It is similar in style to expectation maximization (EM), 
especially  the $\alpha$-EM \citep[e.g.,][]{matsuyama2003alpha} which uses alpha-divergence in EM. However, our method is designed from an optimization perspective for enforcing the label consistency on augmented data, rather than a generative modeling perspective underlying EM.  
%
Empirically, we approximate $\E_{x'\sim \P_x}[\cdot]$ (see Eqn.~\ref{equ:beta_consistency}) by drawing a number of $n$ random samples $\{x_i'\}_{i=1}^n$ from $\P_x$.
See Algorithm~\ref{alg:main} for details.

\begin{algorithm*}[t]
\caption{AlphaMatch: Improving Consistency for SSL with Alpha-divergence}
\label{alg:main}
\begin{algorithmic}[1]
\STATE {\bf Input:} labeled data $\D_s$; unlabeled data $\D_u$; regularization coefficient $\lambda$;  alpha-divergence hyper-parameters $\alpha$ and $\beta$; 
number of augmentation samples $n$;  initial model parameter $\theta_0$. 
\FOR{iteration $t$}
    \STATE Randomly sample a labeled batch $\mathcal B_s$ from $\D_s$ and an unlabeled batch 
    $\mathcal B_u$ from $\D_u$.  
    \FOR{each $x$ in $\mathcal B_u$}
    \STATE Apply data augmentation on $x$ for $n$ times, yielding augmented examples $\{x_i'\}_{i=1}^n$
    \STATE {Fixing} $\theta = \theta_{t}$, 
    update 
    $\gamma(\cdot|x)$ with 
    \begin{align*}
\gamma_{t+1}(\cdot~|~x) \gets \bigg(  (1-\beta) \times p_{\theta_{t}} (\cdot~|~x)^{1-\alpha} +    \frac{\beta}{n} \times \sum_{i=1}^n  \bigg [ p_{\theta_{t}}(\cdot~|~x_i')^{1-\alpha}\bigg]\bigg )^{\frac{1}{1-\alpha}}.
\end{align*}
     \STATE  Approximate $\Psi_{\alpha, \beta}(\theta, \gamma_{t+1}, x)$ in \eqref{equ:beta_consistency} with augmented examples $\{x_i'\}_{i=1}^n$ accordingly. \vspace{.2\baselineskip}
     \ENDFOR
     \STATE  Update $\theta_{t+1}$ by applying one step of gradient descent on \eqref{equ:update_theta} over batch $\mathcal B_u$:
     $$
     \theta_{t+1} \leftarrow \theta_t - \epsilon \nabla_{\theta}\Big(\L(\mathcal B_{s};~ \theta)  + \lambda \E_{x\sim \mathcal B_u}[ \Psi_{\alpha,\beta}(\theta, \gamma_{t+1}, x)]  \Big)
\bigg|_{\theta=\theta_t}.
     $$
     
\ENDFOR
\end{algorithmic}
\end{algorithm*}

\paragraph{Time Cost}
As shown in Algorithm \ref{alg:main}, our proposed algorithm introduces an additional latent variable $\gamma$ which has a closed-form solution.
Therefore, compared to KL divergence, we introduce almost no additional time cost.

\section{Experiments}
%
We test  AlphaMatch on a variety of standard SSL benchmarks (e.g. STL-10, CIFAR-10, CIFAR-100 and SVHN) and compare it with a number of state-of-the-art (SOTA) SSL baselines, including MixMatch \citep{berthelot2019mixmatch}, ReMixMatch \citep{berthelot2019remixmatch} and FixMatch \citep{sohn2020fixmatch}.
We show that AlphaMatch achieves the best performance in all benchmark settings evaluated.

We use the code-base provided in FixMatch \citep{sohn2020fixmatch}\footnote{\url{https://github.com/google-research/fixmatch}} for implementation. 
Throughout our experiments, we simply set $\alpha=1.5$ and
$\beta=0.5$ without tuning; we found the default setting yields the best performance in almost all the cases. We provide comprehensive ablation studies on the impact of different choices of $\alpha$ and $\beta$ in section~\ref{sec:ablation}.

\subsection{STL-10}
We conduct experiments on the challenging SLT-10 dataset.
SLT-10 is a realistic and challenging SSL dataset which 
contains 5,000 labeled images, and 
100,000 unlabeled, which are extracted from a similar but broader distribution of images than labelled data. 
The unlabelled data contains other types of animals (bears, rabbits, etc.) and vehicles (trains, buses, etc.) in addition to the ones in the labeled set.
The distribution shift between the labeled and unlabeled data casts a higher challenge for SSL algorithms, 
and requires us to learn models with stronger generalizability. 
AlphaMatch again shows clear advantages over existing methods in this case. 


\paragraph{Settings}
We preprocess the data and split the labeled images into 5 folds with the same data partition as FixMatch \citep{sohn2020fixmatch}, with each partition containing 1,000 labels. 
Specifically, for each dataset, we use 4,000 labelled data and 100,000 unlabelled data to train and use the remained 1,000 labelled data to test.
Thus, a total of 5 models will be trained and evaluated. 
The final performance is averaged over these 5 individual runs. 

We use the Wide ResNet(WRN)-28-2 and WRN-16-8 model for our method and all the baselines.
We compare AlphaMatch with $\pi$-model, unsupervised data augmentation (UDA), MixMatch, ReMixMatch and FixMatch. All baseline results are produced by 
exactly following the same training setting suggested in \citep{sohn2020fixmatch}.
For our method, we use the default setting of $\alpha=1.5$, $\beta=0.5$ and $n=1$.

\paragraph{Results}
We report the averaged accuracy of all 5 runs in Table~\ref{tab:stl}. 
AlphaMatch significantly outperforms all other baselines in this setting. 
In particular, for WRN-28-2, compared with FixMatch, we improve the accuracy from 89.28\% to 90.36\%.
Our performance is also about 1.9\% higher than ReMixMatch. 
For WRN-16-8, we also improve the baselines with a large margin.

\begin{table}[ht]
\begin{center} 
\begin{tabular}{l|cc}
\hline
Method & WRN-28-2 (\%) & WRN-16-8 (\%) \\
\hline
$\pi$-model & 71.39$\pm$1.21 & 74.92$\pm$1.18 \\
UDA & 86.57$\pm$1.06 & 89.14$\pm$0.73 \\
MixMatch & 85.19$\pm$1.24 & 87.52$\pm$0.79 \\
ReMixMatch & 88.42$\pm$0.78 & 91.07$\pm$0.84 \\
FixMatch & 89.28$\pm$0.63 & 91.35$\pm$0.67\\
\hline
AlphaMatch & \textbf{90.36$\pm$0.75} & \textbf{92.83$\pm$0.86}\\
\hline
\end{tabular}
\end{center}
\vspace{5pt}
\caption{ Testing accuracy on STL-10. All averaged over 5 different folds. Results on two different model architectures, WRN-28-2 and WRN-16-8 is reported.}
\label{tab:stl}
\end{table}

\begin{table*}[h]
    \centering
    \setlength{\tabcolsep}{4pt}
    \begin{tabular}{lcccccc}
    \hline 
        \multirow{2}{*}{Method}& \multicolumn{2}{c}{CIFAR-10} & \multicolumn{2}{c}{SVHN} & \multicolumn{2}{c}{CIFAR-100} \\
         \cmidrule(l{3pt}r{3pt}){2-3}  \cmidrule(l{3pt}r{3pt}){4-5}
        \cmidrule(l{3pt}r{3pt}){6-7}
        & 40 labels & 250 labels  & 40 labels & 250 labels & 400 labels & 2500 labels\\
        \cmidrule(l{3pt}r{3pt}){1-1} \cmidrule(l{3pt}r{3pt}){2-3}  \cmidrule(l{3pt}r{3pt}){4-5}
        \cmidrule(l{3pt}r{3pt}){6-7}
        Pseudo-Labeling & - & 50.22$\pm$0.43 & - & 79.79$\pm$1.09 & - & 42.62$\pm$0.46 \\ 
        UDA & 70.95$\pm$5.93 & 91.18$\pm$1.08 & 47.37$\pm$20.51 & 94.31$\pm$2.70 & 40.72$\pm$0.88 & 66.87$\pm$0.22 \\
        MixMatch & 52.46$\pm$11.50 &   88.95$\pm$0.86 & 57.45$\pm$14.53 & 96.02$\pm$0.23 & 33.39$\pm$1.32 & 60.06$\pm$0.37\\
        ReMixMatch & 80.90$\pm$9.64* & 94.56$\pm$0.05* & 96.64$\pm$0.30* & 97.08$\pm$0.48* & {55.72$\pm$2.06}* & {73.57$\pm$0.31}*\\
        FixMatch & 88.71$\pm$3.35 & 94.93$\pm$0.33 & 92.35$\pm$7.65 & 97.36$\pm$0.64 & 59.79$\pm$2.94* & 74.63$\pm$0.22*\\
        \cmidrule(l{3pt}r{3pt}){1-1} \cmidrule(l{3pt}r{3pt}){2-3}  \cmidrule(l{3pt}r{3pt}){4-5}
        \cmidrule(l{3pt}r{3pt}){6-7}
        AlphaMatch & \bf{91.35$\pm$3.38} & \bf{95.03$\pm$0.29} & \bf{97.03$\pm$0.26} & \bf{97.56$\pm$0.32} & \bf{61.26$\pm$3.13}* & \bf{74.98$\pm$0.27}* \\
        \hline 
    \end{tabular}

    \vspace{5pt}
    \caption{ 
    Testing accuracy (\%) of different methods on CIFAR-10, SVHN, and CIFAR-100. All averaged over 5 different  folds. The `*' indicates the results are achieved by combining the distribution alignment loss proposed in ReMixMatch \citep{berthelot2019remixmatch}. 
    }
    \label{tab:realistic_ssl}
\end{table*}

\subsection{CIFAR-10, SVHN and CIFAR-100}
\label{sec:ssl_cifar_ssl}
We then test AlphaMatch on three widely-used benchmark SSL datasets: 
CIFAR-10 \citep{krizhevsky2009learning}, 
SVHN \citep{netzer2011reading} 
and CIFAR-100. 
With only 4 labeled data per class, 
 we achieve  91.28\%$\pm$3.41\% accurcy on CIFAR-10, 97.03\%$\pm$0.26\% on SVHN, and 61.27\%$\pm$3.13\%  on  CIFAR-100, 
all of which yield 
significantly improvement over prior SSL results on this task.  

\paragraph{Settings} 
For a fair comparison, 
we proceed with minimum changes to the code provided by \citep{sohn2020fixmatch}.
Specifically, we use the same setting and random seeds 
as in FixMatch \citep{sohn2020fixmatch} to 
generate labeled and unlabeled data partitions. 
We use wide ResNet-28-2 \citep{zagoruyko2016wide} with 1.5M parameters as our prediction model,
and then train it using SGD with cosine learning rate decay for 1024 epochs. 
We use CTAugment \citep{berthelot2019remixmatch} for data augmentation as suggested in FixMatch \citep{sohn2020fixmatch}. 

We evaluate two SSL settings, which 
use 4 (resp. 25) labeled images per class, yielding 
40 (resp. 250) labeled images in CIFAR-10 and SVHN  and 
400 (resp. 2500) in  CIFAR-100. 
%
%
The remaining data in the training set is regarded as unlabelled data.
For our method, we set $\alpha=1.5$ and $\beta=0.5$ by default 
and investigate the effect of $\alpha$ and $\beta$ in section \ref{sec:ablation}.
In order to maintain the similar training computation cost as FixMatch, 
We set the number of augmentation applied per image $n=1$ as default; the effect of $n$ is studied in section \ref{sec:ablation}. 
Following FixMatch, we combine AlphaMatch with an additional distribution alignment loss on  
CIFAR-100, which is proposed in ReMixMatch \citep{berthelot2019remixmatch}.

\vspace{-8pt}
\paragraph{Fully Supervised Baselines}
As a reference, 
we train fully-supervised baselines, i.e., training the models
with all the training labels available. 
The test accuracy is 96.14\%$\pm$0.03\%, 97.89\%$\pm$0.02\% and  79.17\%$\pm$0.01\% on CIFAR-10, SVHN and CIFAR-100, respectively. 

\vspace{-8pt}
\paragraph{Results}
We report the test accuracy of all baselines \citep[][]{lee2013pseudo, xie2019uda, berthelot2019mixmatch, berthelot2019remixmatch, sohn2020fixmatch} along with AlphaMatch in Table \ref{tab:realistic_ssl}.
All results are averaged over 5 random trials with different data partitions.
We can see that the models trained with AlphaMatch achieve the best performance in all the settings. 
The gain is especially significant
when the number of labeled data examples is limited (e.g. 4 labeled image per class).
In particular, compared with FixMatch, our method achieves  a 2.64\% improvement on  CIFAR-10 
and 4.58\% on SVHN for test accuracy. 
On CIFAR-100, when combined with the distribution alignment loss as suggested in   \citep{berthelot2019remixmatch},  AlphaMatch yields the best performance, 
achieving
$61.26\%\pm3.13\%$ accuracy with 4 labels per class and 
and  $74.98\%\pm0.27\%$ with 25 labels per class.   
\subsection{Point Cloud Classification}
To further verify the effectiveness of our method, 
we conduct experiments for semi-supervised 3D point cloud classification.
We test a number of prior art SSL baselines and our
algorithm on ModelNet40 \citep{chang2015shapenet} using the SOTA Dynamic graph convolution neural network (DGCNN) \citep{wang2019dynamic}. 

\paragraph{Dataset}
ModelNet40 is the most widely adopted benchmark for point-cloud classification.  
It contains objects from 40 common categories.  There are 9840 objects in the training set and 2468 in the test set. 
In the experiment, we randomly select 100 objects for each category as labelled data and treat the other objects in the training set as unlabelled data.
It means, 
we use 4,000 labelled data and 5,420 unlabelled data during training and then evaluate the performance on the original test set.

\paragraph{Settings}
For all the baselines and our method, we use Gaussian blur $\mathcal{N}(0, 0.02)$ as weak augmentation, and using additional randomized jittering as strong augmentation.
For the DGCNN model, we use 2048 number of particles for each object and set the number of neighbours to 20.
We train the model with 2,00 epochs with SGD with cosing learning rate decay and 0.9 momentum.

\paragraph{Result}
We report the averaged accuracy of all 5 runs in Table~\ref{tab:pointcloud}. 
AlphaMatch significantly outperforms all other baselines in this setting. 
In particular, compared with FixMatch, we improve the accuracy from 86.5\% to 88.3\%.
Compared to other baselines, the proposed methods can also boost the performance by a large margin.
The result shows that our method  can also be applied to other settings except image classification.

\begin{table}[h]
\begin{center} 
\begin{tabular}{l|c}
\hline
Method & ModelNet40 Accuracy (\%) \\
\hline
$\pi$-model & 81.82$\pm$1.18 \\
UDA & 86.15$\pm$1.13  \\
MixMatch & 85.34$\pm$1.05  \\
ReMixMatch & 85.66$\pm$0.92  \\
FixMatch & 86.47$\pm$0.79 \\
\hline
AlphaMatch & \textbf{88.32$\pm$0.84} \\
\hline
\end{tabular}
\end{center}
\vspace{5pt}
\caption{ Testing accuracy on ModelNet40. All averaged over 5 different folds.}
\label{tab:pointcloud}
\end{table}

\subsection{Ablation Studies}
\label{sec:ablation}

\begin{figure*}[t]
\centering
\begin{tabular}{cc}
\raisebox{2.0em}{\rotatebox{90}{  Test Accuracy}}
\includegraphics[height=0.19\textwidth]{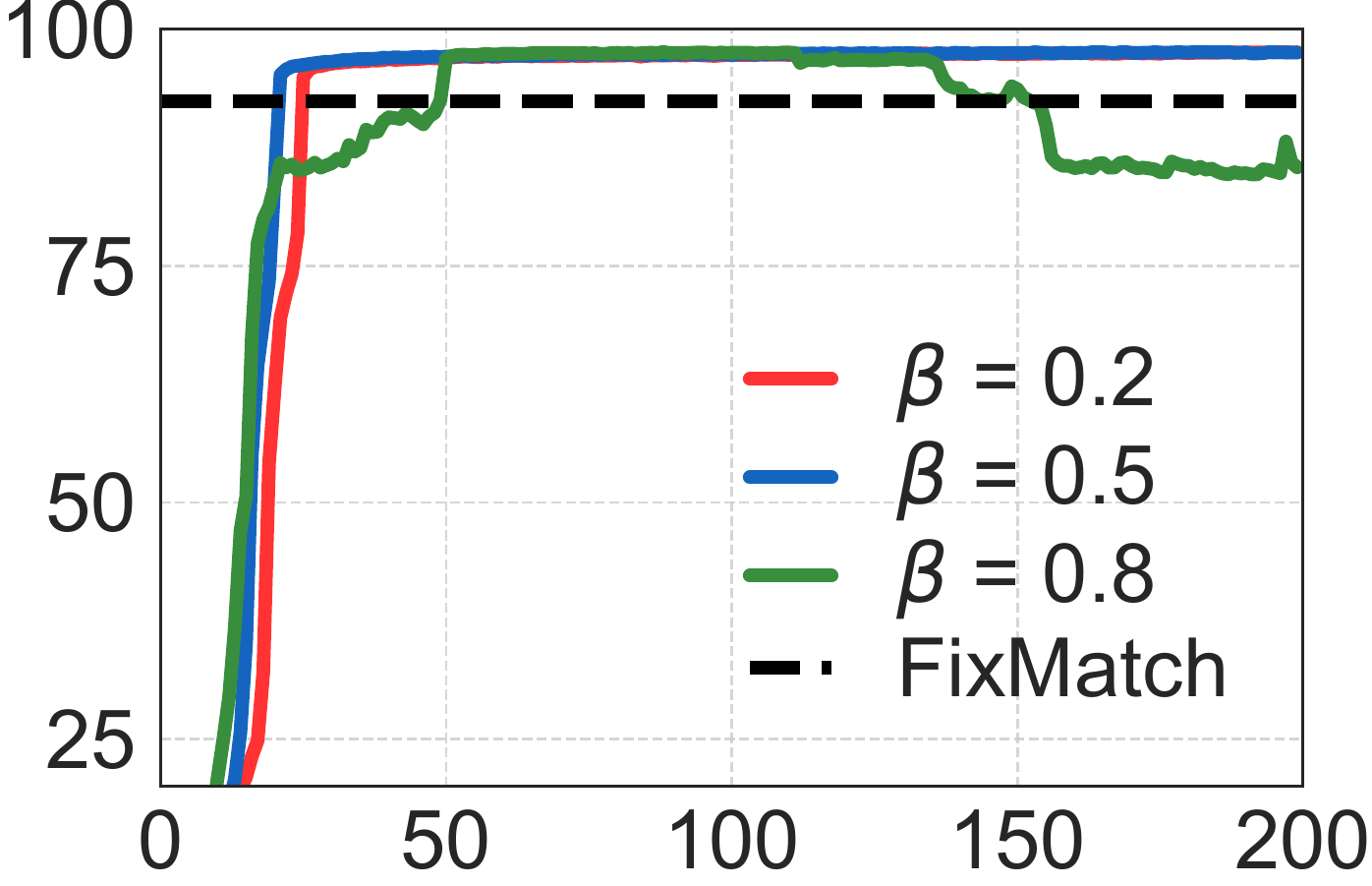} &
\raisebox{2.0em}{\rotatebox{90}{ Test Accuracy}}
\includegraphics[height=0.19\textwidth]{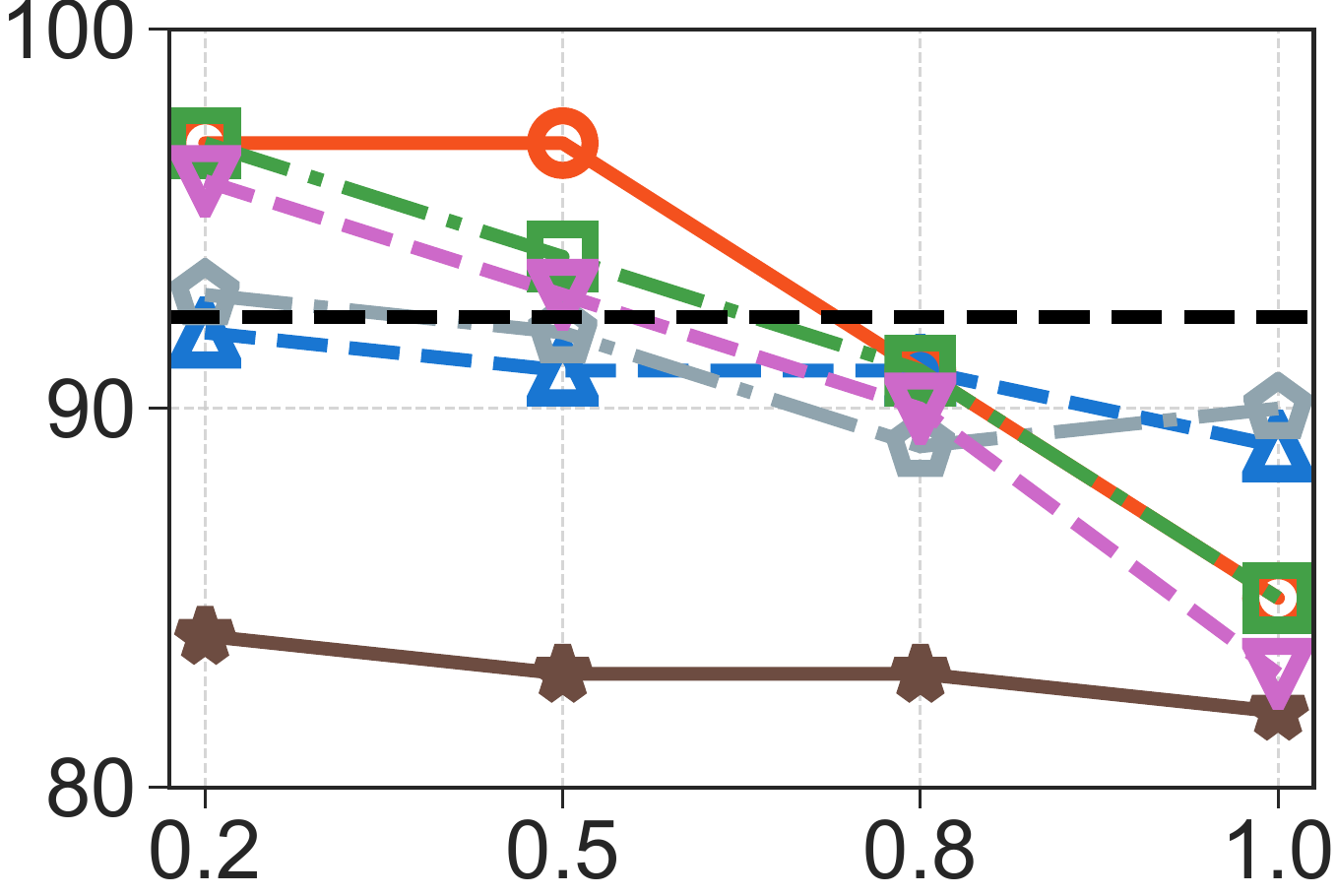} 
\raisebox{0.5em}{\includegraphics[width=0.145\textwidth]{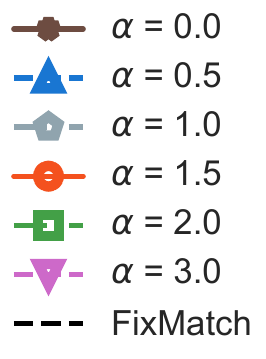}}   \\
 ~~~(a) Training epoch ($\alpha=1.5$) & (b) choices of $\beta$ ~~~~~~~~~~~~~~~  \\
\end{tabular}
\vspace{5pt}
\caption{(b) The Learning curve of AlphaMatch on SVHN, when using  $\alpha=1.5$ and varying $\beta$.  The x-axis denotes the training epoch  and y-axis represents the test accuracy. 
The dashed line shows the result of FixMatch. 
(b) Testing accuracy of AlphaMatch with various $\alpha$ and $\beta$ on the SVHN dataset. 
All models are trained with 4 labeled examples per class. 
} 
\label{fig:ablation}
\end{figure*}
\vspace{5pt}

\begin{table*}[ht]
\begin{center} 
\begin{tabular}{c|cccc}
\hline 
$n$ augmented examples & 1 & 2 & 4 & 10\\
\hline
CIFAR-10 Accuracy (\%) & 91.35$\pm$3.38 & 91.48$\pm$3.41 & 91.63$\pm$3.12 & 91.59$\pm$3.08 \\
\hline
SVHN Accuracy (\%) & 97.03$\pm$0.26 & 97.07$\pm$0.19 & 97.18$\pm$0.23 & 97.18$\pm$0.21 \\
\hline 
\end{tabular}
\end{center}
\vspace{5pt}
\caption{ Comparison on testing accuracy for various $n$. All results are averaged over 5 random trials with the same settings as section \ref{sec:ssl_cifar_ssl}.
}
\label{tab:ablation_n}
\end{table*}

\paragraph{Impact of $\alpha$ and $\beta$} 
We follow all training settings as Section~\ref{sec:ssl_cifar_ssl}.
 All models are trained with 4 labeled examples per class.
On this dataset, FixMatch achieves $92.4\%$ test accuracy (see the dashed black line).
In Figure~\ref{fig:ablation} (a), 
we fix $\alpha=1.5$ and study the impact of $\beta$. 
We find a smaller $\beta$ (e.g. $\beta=0.2$ or $0.5$) often yields more stabilized training; in the contrary, a larger $\beta$ (e.g. $\beta=0.8$) focuses less on clean (or weakly augmented) data and performs worse than small $\beta$ in general. 

We plot in Figure~\ref{fig:ablation} (b) the testing accuracy of AlphaMatch with different $\alpha$ and $\beta $ 
on SVHN dataset.  
As we can see from Figure~\ref{fig:ablation},
a larger $\alpha$ (e.g. $\alpha\ge 1.5$) and a smaller $\beta$ (e.g. $\beta\le 0.5$)
normally achieves better performance than FixMatch. 
This is expected, as using alpha-divergence with large $\alpha$ values 
help propagate high confidence labels and a smaller $\beta$ helps to stabilize the training.
On the other hand, if $\alpha$ is too large, 
the performance may diminish because numerical instability increases.

We observe that $\alpha=1.5$ and $\beta=0.5$ yields the best performance, 
achieving a good balance between consistency regularization and training stability. 

\paragraph{Impact of the number of augmented examples $n$}
In our algorithm, $n$ controls on how many augmented examples is generated to approximate $\P_x$ (see Algorithm \ref{alg:main}). 
We can expect that $\{x_i'\}_{i=1}^n$ forms increasingly better approximation for $\P_x$ with a larger $n$. 
In this section, 
we perform an in-depth analysis on the effect of different $n$ values.
Intuitively, a smaller $n$ leads to better energy-efficiency
while a larger $n$ is more computationally expensive but may yield more robust estimation hence produce better performance.

We test AlphaMatch on CIFAR-10 and SVHN, with the same settings as section~\ref{sec:ssl_cifar_ssl}. 
Table~\ref{tab:ablation_n} shows the performance of various $n$.
All results are averaged over 5 random trials.
We find $n=1$ often performs competitively and     
a larger $n$ (e.g. $n=4$, $n=10$) yields slightly improvements in general. 
However, a large $n$ can significantly increase the computation cost
since it increases the forward and backward time cost of the training model, and more critically, it requires a larger GPU memory cost.
Considering its huge time/memory cost and mild improvement, we recommend to use $n=1$ in practice.

\section{Related Works}  


Semi-supervised learning (SSL) has been a classical subfield of machine learning with a large literature. 
Examples of classical methods include  
 transductive models \citep[e.g.][]{gammerman2013learning, demiriz1999semi, joachims1999transductive},
 co-training \citep[e.g.][]{blum1998combining,  nigam2000analyzing},  entropy minimization \citep[e.g.][]{grandvalet2005semi, lee2007learning},  graph-based models \citep[e.g.][]{blum2001learning, he2018amc, zhu2002learning, zhou2004learning,  belkin2004regularization, wang2007label, wu2007transductive}
and many more (see \citep[e.g.][]{zhu2005semisurvey, chapelle2009semi}). 

Due to the recent success of deep learning,
combining generative model \cite{dai2017good, higgins2016beta} and adversarial training \cite{miyato2015vat} with SSL achieves meaningful improvement over the classic SSL methods in many real-world problems, e.g. image classification \cite{xie2019uda, wang2020attentivenas}, segmentation \cite{papandreou2015weakly}, detection \cite{jeong2019consistency}.
Most recently, data augmentation has been introduced into SSL and achieved great success.
$\Pi$-model, for example,  shares the similar idea of  \eqref{eq:ssl-loss}, but replaces the KL divergence with L2 distance and replace $x$ in \eqref{eq:ssl-loss} with another random copy of augmentation $x''$. 
MixMatch \cite{berthelot2019mixmatch} and ReMixMatch \cite{berthelot2019remixmatch}, 
use \emph{mixup} \cite{zhang2017mixup} to do data augmentation.
Based on MixMatch, 
ReMixMatch uses some additional loss and new data augmentation method to improve the performance.
These methods force the label consistency between weakly augmented examples (or examples without augmentation) and strong augmented examples, and improve the state-of-the-art results on classification tasks by a large margin.

Closely related to our work, 
a series of recent SSL methods have been proposed based on the general idea of 
enforcing the label prediction of an image to be consistent with its augmented counterparts. 
The label consistency has been mostly measured by either KL divergence 
 \citep[e.g.,][]{xie2019uda,berthelot2019mixmatch, berthelot2019remixmatch, miyato2015vat, sohn2020fixmatch}, 
 or mean squared error \citep[e.g.,][]{laine2016temporal, tarvainen2017mean,sajjadi2016regularization,rasmus2015pi}. 
 Almost all these methods use iterative regularization processes similar to \eqref{eq:ssl-loss}-\eqref{equ:udagrad}, 
 with the stop-gradient trick (see Section \ref{sec:bg}). 
Compared to these works,  
our method first introduces alpha-divergence as consistency measure,  
and equip it with a convergent EM-like matching algorithm to achieve better results. 


Most recently, \citet{chen2020exploring} presents a similar EM-like approach for self-supervised learning.
For each step, \citet{chen2020exploring} first closed-form optimizes a local latent variable (similar to our $\gamma$) and then updates the model parameters with a fixed $\gamma$.
This motivates us to further explore our method to more general topics.

\section{Conclusion}
In this paper, 
we propose to use alpha-divergence and a new optimization-based framework to build a semi-supervised learning algorithm based on data augmentation.
The proposed AlphaMatch is simple yet powerful. 
With only a few lines of extra code to implement alpha-divergence and the EM-like update,
it achieves the state-of-the-art performance on various benchmarks. 

For future work, we will apply our algorithm to more practical tasks, e.g. 2D/3D segmentation, machine translation, object detection. 
By further extending and improving our framework, it is promising to push the frontier of ML with limited data to close the gap between few-shot learning and SSL.
Finally, we plan to extend our algorithm to unsupervised learning, out-of-distribution detection and other related topics.


{\small
\bibliography{ref}
\bibliographystyle{nips2018}
}


\end{document}